\documentclass[11pt]{article}

\usepackage[utf8]{inputenc}
\usepackage[T1]{fontenc}
\usepackage{lmodern}

\usepackage[margin=1in]{geometry} 
\usepackage{microtype}            

\usepackage{amsmath,amssymb,amsfonts,mathtools,bm}
\allowdisplaybreaks 

\usepackage{algorithm}
\usepackage{algpseudocode}

\usepackage{graphicx}
\usepackage{booktabs}

\usepackage[hidelinks]{hyperref}

\usepackage{xspace}

\DeclareMathOperator{\Tr}{Tr}

\newcommand{\B}[1]{\bm{#1}}

\title{\vspace{-0.5em}Fast Swap-Based Element Selection for Multiplication-Free Dimension Reduction}
\author{Nobutaka Ono\thanks{Tokyo Metropolitan University, 6--6 Asahigaoka, Hino-shi, Tokyo 191--0065, Japan. \texttt{onono@tmu.ac.jp}}}
\date{} 

\begin{document}
\maketitle

\begin{abstract}
\noindent
In this paper, we propose a fast algorithm for \emph{element selection}, a multiplication-free form of dimension reduction that produces a dimension-reduced vector by simply selecting a subset of elements from the input.
Dimension reduction is a fundamental technique for reducing unnecessary model parameters, mitigating overfitting, and accelerating training and inference.
A standard approach is principal component analysis (PCA), but PCA relies on matrix multiplications; on resource-constrained systems, the multiplication count itself can become a bottleneck.
Element selection eliminates this cost because the reduction consists only of selecting elements, and thus the key challenge is to determine which elements should be retained.
We evaluate a candidate subset through the minimum mean-squared error of linear regression that predicts a target vector from the selected elements, where the target may be, for example, a one-hot label vector in classification.
When an explicit target is unavailable, the input itself can be used as the target, yielding a reconstruction-based criterion.
The resulting optimization is combinatorial, and exhaustive search is impractical.
To address this, we derive an efficient formula for the objective change caused by swapping a selected and an unselected element, using the matrix inversion lemma, and we perform a swap-based local search that repeatedly applies objective-decreasing swaps until no further improvement is possible.
Experiments on MNIST handwritten-digit images demonstrate the effectiveness of the proposed method.
\end{abstract}


\section{Introduction}
Recent advances in machine learning have increased both the volume of data and the dimensionality of each datum. To reduce computational cost and improve generalization, dimension reduction is widely used as preprocessing~\cite{Cunningham2015}. Principal component analysis (PCA) is a basic approach~\cite{Pearson1901,Eckart1936,Jolliffe2002}, but its projection requires many multiplications, which can be costly on resource-limited devices such as hearing aids and embedded systems~\cite{Sze2017}. Random projections provide another simple reduction method~\cite{JohnsonLindenstrauss1984,Achlioptas2003}, yet they also require multiplications. Feature/variable selection has also been extensively studied as an alternative to projection-based reduction~\cite{GuyonElisseeff2003,KohaviJohn1997}; however, many existing methods are designed primarily for prediction accuracy with a given learner rather than for regression fidelity under an explicit inference-cost constraint.

In this study, we take the \emph{element selection} viewpoint to realize multiplication-free dimension reduction: the dimension-reduced vector is obtained by extracting a fixed subset of input elements, which requires no scalar multiplications (i.e., only memory access).
The key remaining issue is how to determine a good subset under a principled criterion and with practical offline computation time.

\paragraph{Related work and motivation.}
Element selection for multiplication-free dimension reduction has already been used as a preprocessing step in low-complexity audio machine learning systems, such as DNN-based speech enhancement for hearing aids~\cite{HarutaOno2021EUSIPCO} and CNN-based musical instrument classification~\cite{KatoEtAl2024IWAENC}.
These studies optimize selected indices offline and then deploy only the selection at runtime, so the selection algorithm itself must be fast enough to be used repeatedly during system design and hyper-parameter exploration.
Meanwhile, more general frameworks based on nonconvex sparse optimization have been proposed to handle a wider class of loss functions for restorability-based element selection~\cite{KawamuraEtAl2023ICASSP,KawamuraEtAl2024IEEEAccess}, at the cost of requiring many iterations.

\paragraph{Contribution.}
We formulate element selection as a discrete optimization that minimizes the mean-squared linear regression error from selected elements to a target vector, which includes reconstruction as a special case.
We propose a swap-based local search and derive an accelerated evaluation of candidate swaps using the matrix inversion lemma, reducing the dominant computations to a small $2\times 2$ matrix inversion per candidate.

\section{Problem Formulation}
\subsection{Element Selection as Multiplication-Free Dimension Reduction}
Let $\B{x}$ denote an $N$-dimensional datum, and consider obtaining, by dimension reduction, a $K$-dimensional (with $K<N$) dimension-reduced vector $\B{y}$ from $\B{x}$. In a linear dimension reduction method, multiplying by a $K\times N$ matrix $P$ yields
\begin{equation}
  \B{y} = P \B{x},
  \label{eq:linear}
\end{equation}
as the dimension-reduced vector $\B{y}$.
In general, $P$ is taken to be, for example, a projection matrix onto principal components. However, computing $P\B{x}$ requires $KN$ scalar multiplications.
For instance, reducing 1000 dimensions to 100 dimensions requires $10^5$ multiplications. Since dimension reduction must be performed not only during training but also at test time in machine learning, on embedded devices and the like, the computation for this dimension reduction itself can become a substantial burden.

To avoid this, we extract $K$ elements out of the $N$ elements of $\B{x}$ and compute
\begin{equation}
  \B{y} = (x_{\sigma(1)}\ x_{\sigma(2)}\ \cdots\ x_{\sigma(K)})^\top
\end{equation}
as the dimension-reduced vector $\B{y}$,
where $\{\sigma(i)\mid i=1,\cdots,K\}$ denotes mutually distinct element indices chosen from $1$ to $N$.
Clearly, the number of multiplications required to obtain $\B{y}$ from $\B{x}$ is $0$.

This element selection is equivalent to choosing, as $P$ in eq. (\ref{eq:linear}), a matrix whose $(i,j)$-th component is
\begin{equation}
  p_{ij}=
  \begin{cases}
    1 & \text{if } j=\sigma(i),\\
    0 & \text{otherwise},
  \end{cases}
  \label{eq:select_mrx}
\end{equation}
that is, a matrix in which each row vector contains exactly one $1$ and all other entries are $0$.
Therefore, element selection can be regarded as one form of linear dimension reduction.

\subsection{Derivation of the Objective Function}

Here, the goal of element selection is to reduce the dimensionality of an input
vector for a downstream machine-learning model. Let $\B{x}\in\mathbb{R}^N$
denote the original input, and let $\B{z}\in\mathbb{R}^M$ denote the target
vector, i.e., the model output of interest (e.g., a one-hot label vector in
classification). When a clear target vector cannot be defined, we may simply
set $\B{z}=\B{x}$, which corresponds to evaluating reconstruction performance.
To assess the quality of the dimension-reduced vector $\B{y}=P\B{x}$, we consider
estimating $\B{z}$ linearly from $\B{y}$ as
\begin{equation}
  \hat{\B{z}} = Q \B{y},
\end{equation}
where $Q\in\mathbb{R}^{M\times K}$.
We evaluate the quality of the dimension-reduced vector by the mean squared regression
error
\begin{align}
  E(P)
  &= \min_{Q} E\big[\|\hat{\B{z}}-\B{z}\|^2\big]
   = \min_{Q} E\big[\|Q\B{y}-\B{z}\|^2\big].
\end{align}
The minimizer is
\begin{equation}
  Q_0 = E[\B{z}\B{y}^\top]\, E[\B{y}\B{y}^\top]^{-1}.
  \label{eq:Q0}
\end{equation}
Let
\begin{align}
  V^{(xx)} &= E[\B{x}\B{x}^\top], \\
  V^{(xz)} &= E[\B{x}\B{z}^\top], \\
  V^{(zz)} &= E[\B{z}\B{z}^\top].
\end{align}
Using $\|\B{z}\|^2=\Tr(\B{z}\B{z}^\top)$ and cyclicity of the trace, we obtain
\begin{align}
  E(P)
  &= \Tr(V^{(zz)})
   - \Tr\!\big(
        E[\B{z}\B{y}^\top]\,
        E[\B{y}\B{y}^\top]^{-1}\,
        E[\B{y}\B{z}^\top]
      \big).
\end{align}
Since $\B{y}=P\B{x}$, we have
\begin{align}
  E[\B{y}\B{y}^\top] &= P V^{(xx)} P^\top,\\
  E[\B{z}\B{y}^\top] &= V^{(zx)} P^\top,\\
  E[\B{y}\B{z}^\top] &= P V^{(xz)},
\end{align}
where $V^{(zx)} = (V^{(xz)})^\top$.
Therefore,
\begin{align}
  E(P)
  &= \Tr(V^{(zz)})
   - \Tr\!\big(
       V^{(zx)} P^\top (P V^{(xx)} P^\top)^{-1} P V^{(xz)}
     \big).
\end{align}
Since $\Tr(V^{(zz)})$ is constant with respect to $P$, we define the objective
\begin{equation}
  J(P)
  = \Tr\!\big(
      V^{(zx)} P^\top (P V^{(xx)} P^\top)^{-1} P V^{(xz)}
    \big),
  \label{eq:J_reg}
\end{equation}
and maximize $J(P)$ over the element index set
$I = \{\sigma(i)\mid i=1,\ldots,K\}$, where $P$ is determined by $I$ as
in~\eqref{eq:select_mrx}.
This is a discrete optimization problem.

\section{Fast Swap-Based Optimization}

\subsection{Exhaustive Search}
The most direct solution is to compute \eqref{eq:J_reg} for all $\binom{N}{K}$ selections and find $I$ maximizing it.
Using Stirling's approximation
\begin{equation}
  n! \sim \sqrt{2\pi n}\left(\frac{n}{e}\right)^{\!n},
  \label{eq:stirling}
\end{equation}
we estimate
\begin{equation}
  \binom{N}{K} \sim
  \sqrt{\frac{N}{2\pi K(N-K)}}\cdot
  \frac{N^N}{K^K (N-K)^{N-K}},
  \label{eq:binom}
\end{equation}
which is intractable, e.g.\ $>10^{139}$ for $N{=}1000$, $K{=}100$.

\subsection{Objective Reformulation for Efficient Evaluation}

To simplify the discussion, let $\sigma(1),\ldots,\sigma(N)$ be a permutation of $\{1,\ldots,N\}$,
and consider the permuted vector $(x_{\sigma(1)},\ldots,x_{\sigma(N)})^\top$ and its covariance
\begin{equation}
  \tilde V^{(xx)} =
  \begin{pmatrix}
    A      & B_x \\
    B_x^\top & C
  \end{pmatrix},
  \label{eq:Vtilde_block}
\end{equation}
where
\begin{align}
  \tilde V^{(xx)}_{ij} &= V^{(xx)}_{\sigma(i),\sigma(j)}, \label{eq:Vtilde_comp} \\
  A_{ij} &= \tilde V^{(xx)}_{ij} \quad (1\le i,j\le K), \\
  (B_x)_{ij} &= \tilde V^{(xx)}_{i,K+j} \quad (1\le i\le K,\;1\le j\le N-K).
\end{align}
Thus $A\in\mathbb{R}^{K\times K}$ is the covariance of the selected elements
$\{x_{\sigma(1)},\ldots,x_{\sigma(K)}\}$.

Next, let $\B{z}$ denote the $M$-dimensional target vector introduced in
Section~2.2, and define the cross-covariance between $\B{x}$ and $\B{z}$ as
\begin{equation}
  V^{(xz)} = E[\B{x}\B{z}^\top]\in\mathbb{R}^{N\times M}.
\end{equation}
Under the same permutation of $\B{x}$, we partition
\begin{equation}
  \tilde V^{(xz)} =
  \begin{pmatrix}
    B \\
    D
  \end{pmatrix}
  =
  \begin{pmatrix}
    V^{(xz)}_{\sigma(1),:} \\
    \vdots \\
    V^{(xz)}_{\sigma(K),:} \\
    \hline
    V^{(xz)}_{\sigma(K+1),:} \\
    \vdots \\
    V^{(xz)}_{\sigma(N),:}
  \end{pmatrix},
\end{equation}
so that $B\in\mathbb{R}^{K\times M}$ is the cross-covariance between the selected
elements and the target,
\begin{equation}
  B = E[\B{y}\B{z}^\top],
\end{equation}
where
$\B{y}=(x_{\sigma(1)},\ldots,x_{\sigma(K)})^\top$ is the $K$-dimensional dimension-reduced vector obtained by element selection.

The covariance of $\B{y}$ is $A=E[\B{y}\B{y}^\top]$ and the cross-covariance
between $\B{y}$ and $\B{z}$ is $B$. Therefore, the objective \eqref{eq:J_reg} can be rewritten as
\begin{equation}
  J(P)
  = \Tr\!\bigl(B^\top A^{-1} B\bigr).
  \label{eq:J_block}
\end{equation}


\subsection{Fast Swap-Based Element Selection}
Here, we consider a swap-based local improvement (cf. 2-opt heuristic~\cite{Croes1958}). Specifically, we consider an iterative procedure in which one element currently selected and one element not selected are swapped, and the swap is accepted only when it leads to an increase in the objective function. This process is repeated until no further improving swap exists. This yields a local optimum of \eqref{eq:J_reg}.

Let the selected set be $I=\{\sigma(1),\ldots,\sigma(K)\}$ and the complement
$I^{c}=\{\sigma(K+1),\ldots,\sigma(N)\}$.
Pick $i\in I$ and $j\in I^{c}$, tentatively swap $\sigma(i)$ and $\sigma(j)$,
accept if $J$ increases, otherwise revert. For each $i$, try all
$j$ and accept the best improving swap; repeat until no improvement. This
yields a local optimum of \eqref{eq:J_block}.

In \eqref{eq:J_block}, inverting $A$ dominates the cost. When swapping
$\sigma(i)$ and $\sigma(j)$ with $1\le i\le K<j\le N$, only the $i$-th row and
column of $A$ and the $i$-th row of $B$ change.

From the permuted covariance $\tilde V^{(xx)}$, the update of $A$ is the rank-2 update:
\begin{equation}
  A \leftarrow A + e_i f^\top + f e_i^\top,
  \label{eq:A_update}
\end{equation}
where $e_i$ is the $i$-th standard basis vector in $\mathbb{R}^{K}$ and
$f\in\mathbb{R}^{K}$ is defined by
\begin{equation}
  f_n =
  \begin{cases}
    \tilde V^{(xx)}_{n j} - \tilde V^{(xx)}_{n i}, & n\neq i,\\[0.25em]
    \bigl(\tilde V^{(xx)}_{j j}-\tilde V^{(xx)}_{i i}\bigr)/2, & n=i.
  \end{cases}
  \label{eq:f_def}
\end{equation}
On the other hand, the cross-covariance $B$ is updated only in the $i$-th row as
\begin{equation}
  B \leftarrow B + e_i h^\top,
  \label{eq:B_update}
\end{equation}
where $h\in\mathbb{R}^{M}$ is the difference between the $j$-th and $i$-th rows
of $\tilde V^{(xz)}$:
\begin{equation}
  h = (\tilde V^{(xz)}_{j,:})^\top - (\tilde V^{(xz)}_{i,:})^\top.
  \label{eq:h_def}
\end{equation}

Let
\begin{align}
  F &= (\,e_i\ \ f\,), \\
  X &=
  \begin{pmatrix}
    0 & 1\\
    1 & 0
  \end{pmatrix}.
\end{align}
Then \eqref{eq:A_update} can be written compactly as
\begin{equation}
  A \leftarrow A + F X F^\top.
\end{equation}
Let $W=A^{-1}$. By the matrix inversion lemma,
\begin{align}
  (A + F X F^\top)^{-1}
  &= W \;+\; W F\, G^{-1} F^\top\! W,
\end{align}
where
\begin{equation}
  G = -\bigl(X + F^\top W F\bigr)\in\mathbb{R}^{2\times 2}.
\end{equation}
Since $G$ is $2\times 2$, the inverse $G^{-1}$ can be computed cheaply.

Because $J=\Tr(B^\top W B)$, the change in the objective caused by
swapping $\sigma(i)$ and $\sigma(j)$ is
\begin{align}
  \Delta J_{ij}
  &= \Tr\!\Bigl[
      (B + e_i h^\top)^\top
      (W + W F G^{-1} F^\top W)
      (B + e_i h^\top)
      - B^\top W B
    \Bigr].
  \label{eq:deltaJ1}
\end{align}
To reduce the computational cost of \eqref{eq:deltaJ1}, we use cyclic
reordering of the trace and introduce the following auxiliary matrices:
\begin{align}
  R &= W F \in\mathbb{R}^{K\times 2}, \\
  S &= B^\top R \in\mathbb{R}^{M\times 2}, \\
  t &= B h \in\mathbb{R}^{K}, \\
  u &= R^\top t \in\mathbb{R}^{2}.
\end{align}
Furthermore, let $w_i$ denote the $i$-th column of $W$ and let $r_i$ denote the
$i$-th column of $R^\top$ (equivalently, the transpose of the $i$-th row of $R$).

Then \eqref{eq:deltaJ1} can be rearranged as
\begin{align}
  \Delta J_{ij}
  &= \Tr\!\bigl( G^{-1} S^\top S \bigr)
   + 2\bigl( t^\top w_i + u^\top G^{-1} r_i \bigr)
   + \bigl( W_{ii} + r_i^\top G^{-1} r_i \bigr)\, h^\top h.
  \label{eq:deltaJ2}
\end{align}
All matrix inversions in the inner loop are thus reduced to the $2\times 2$
matrix $G$, which substantially accelerates the evaluation of $\Delta J_{ij}$.

The resulting fast swap-based element selection procedure is summarized in Algorithm~\ref{alg:fast_en}.

\paragraph{Implementation note.} In practice, it is convenient to keep the currently selected elements in the first $K$ positions by explicitly swapping the corresponding rows/columns of $V^{(xx)}$ and the corresponding rows of $V^{(xz)}$ after each accepted swap (as in our MATLAB implementation). Then $A=V^{(xx)}_{1{:}K,1{:}K}$ and $B=V^{(xz)}_{1{:}K,:}$ can be extracted as leading blocks without additional indexing overhead.

\begin{algorithm}[t]
\caption{Fast swap-based element selection with accelerated evaluation}
\label{alg:fast_en}
\begin{algorithmic}[1]
\Require Covariance $V^{(xx)}$ and cross-covariance $V^{(xz)}$, target dimension $K$
\Ensure $I=\{\sigma(1),\ldots,\sigma(K)\}$
\State Initialize a permutation $\sigma(1),\ldots,\sigma(N)$ (e.g., random or identity)
\State Compute $A$ and $B$ from $V^{(xx)}$ and $V^{(xz)}$ (Section~3.2); set $W\gets A^{-1}$
\Repeat
  \State $c_i\gets 0$ for all $i=1,\ldots,K$
  \For{$i=1$ to $K$}
    \State Recompute $A$ and $B$ from $V^{(xx)}$ and $V^{(xz)}$; set $W\gets A^{-1}$
    \For{$j=K{+}1$ to $N$}
      \State Compute $\Delta J_{ij}$ by \eqref{eq:deltaJ2}
    \EndFor
    \State $j^\ast\gets \arg\max_j \Delta J_{ij}$;\quad $\Delta J^\ast\gets \max_j \Delta J_{ij}$
    \If{$\Delta J^\ast>0$}
      \State Swap $\sigma(i)$ and $\sigma(j^\ast)$;\quad $c_i\gets 1$
    \EndIf
  \EndFor
\Until{$\sum_{i=1}^{K} c_i = 0$}
\end{algorithmic}
\end{algorithm}

\section{Experiments}
\subsection{Experimental Conditions}
We used the MNIST handwritten digit database~\cite{LeCun1998}. From the 70{,}000 images of size $28{\times}28$ pixels, we used the 60{,}000 training images. Each image was vectorized into a 784-dimensional vector; after subtracting the global mean to make the mean zero, we estimated the covariance matrix. To avoid numerical instability, the covariance matrix was regularized by adding the identity matrix multiplied by $10^{-5}$ times the largest eigenvalue. We set $N{=}784$ and $K{=}100$, and selected the optimal 100 dimensions.
In the experiments of this paper, we simply set the target vector equal to the input, $\B{z}=\B{x}$. In this case, the regression objective \eqref{eq:J_reg} represents the reconstruction loss. 

\subsection{Covariance Structure Before and After Selection}
\begin{figure}[t]
  \begin{center}
    \includegraphics[width=0.44\linewidth]{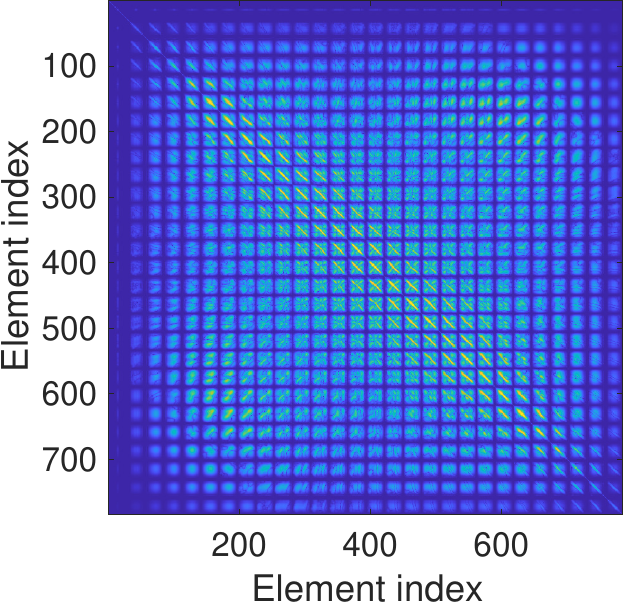}\hfil
    \includegraphics[width=0.44\linewidth]{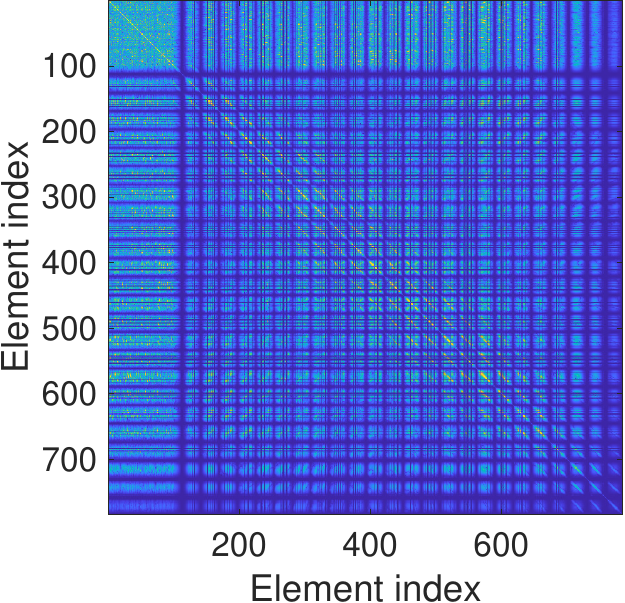}
    \vspace{-5mm}
  \end{center}
  \caption{Covariance matrices before (left) and after (right) element selection. For display, each entry is transformed as $|\cdot|^{0.3}$.}
  \label{fig:cov}
\end{figure}

Figure~\ref{fig:cov} shows the covariance matrices. The left panel is the original covariance; the right is after repeatedly applying element selection and gathering the selected elements as indices $1$ through $100$. For visibility, we raised the absolute value of each entry to the power $0.3$ to compress dynamic range.

\subsection{Change in Normalized Loss}
We plot the normalized regression loss
\begin{align}
  \tilde{E}(P)
  &= \frac{\Tr(V^{(zz)})-J(P)}{\Tr(V^{(zz)})}
   \;=\; 1-\frac{J(P)}{\Tr(V^{(zz)})},
  \label{eq:Enorm}
\end{align}
which coincides with the normalized reconstruction loss in our experiments because $\B{z}=\B{x}$ and hence $V^{(zz)}=V^{(xx)}$.
The loss decreases monotonically and converges; at convergence it was $0.1367$, while PCA with the largest $100$ components gave $0.0856$.

\begin{figure}[t]
  \begin{center}
    \includegraphics[width=0.6\linewidth]{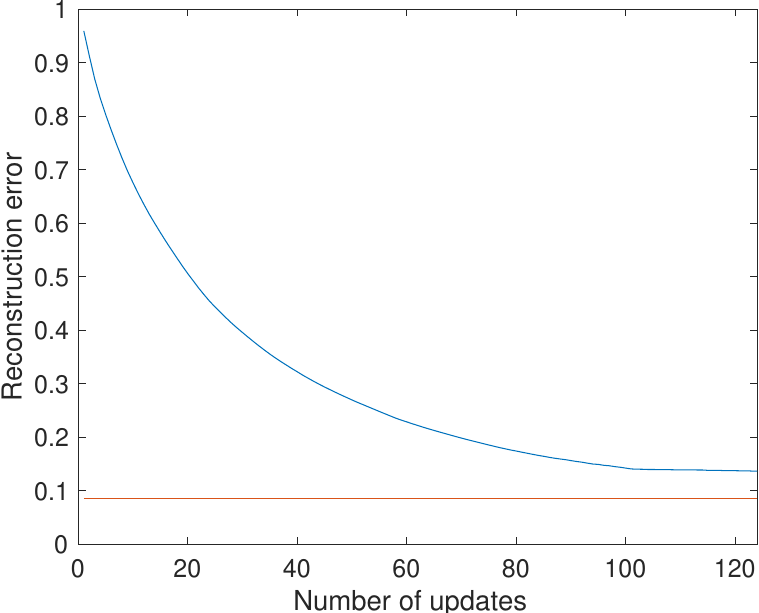}
    \vspace{-5mm}
  \end{center}
  \caption{Change in normalized loss (reconstruction loss in our setting). The horizontal line indicates the loss of PCA.}
  \label{fig:err}
\end{figure}

\subsection{Computation-Time Comparison}
On a notebook PC with an Intel Core i7-7500 (2.7\,GHz) using MATLAB R2019b, a naive implementation that directly recomputes the objective for each candidate swap took about 1{,}800 seconds to converge, whereas the proposed accelerated evaluation (Algorithm~\ref{alg:fast_en}) took about 22 seconds. With further vectorization, the time was about 8 seconds (over $200\times$ speed-up).

\subsection{Visualization of Selected Elements}
\begin{figure}[t]
  \begin{center}
    \includegraphics[width=0.42\linewidth]{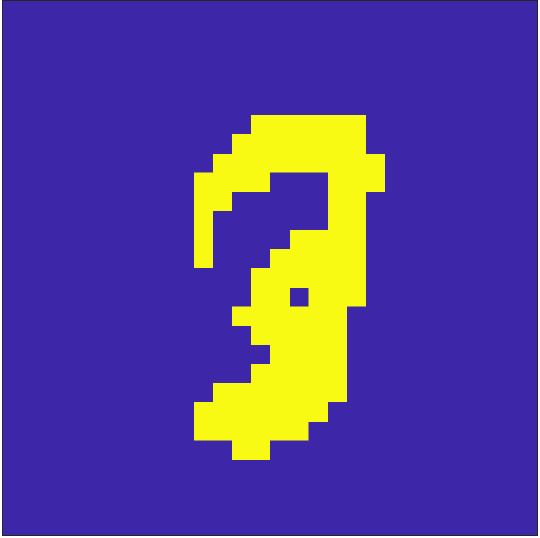}\hfil
    \includegraphics[width=0.42\linewidth]{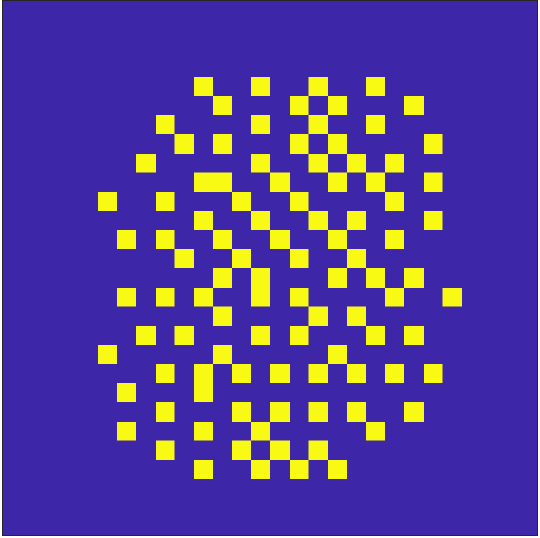}
  \end{center}
  \vspace{-2mm}
  \caption{Pixels selected by variance (left) and by the proposed method (right).}
  \label{fig:pixel}
\end{figure}

Figure~\ref{fig:pixel} compares (i) the 100 pixels with the largest marginal variances and (ii) the 100 pixels selected by the proposed method. Simply choosing high-variance pixels concentrates around the image center, while our method considers inter-pixel correlation and selects a wider spread.

\subsection{Image Reconstruction}
\begin{figure}[t]
  \centering
  \includegraphics[width=\linewidth,trim=1.6cm 0 1.2cm 0,clip]{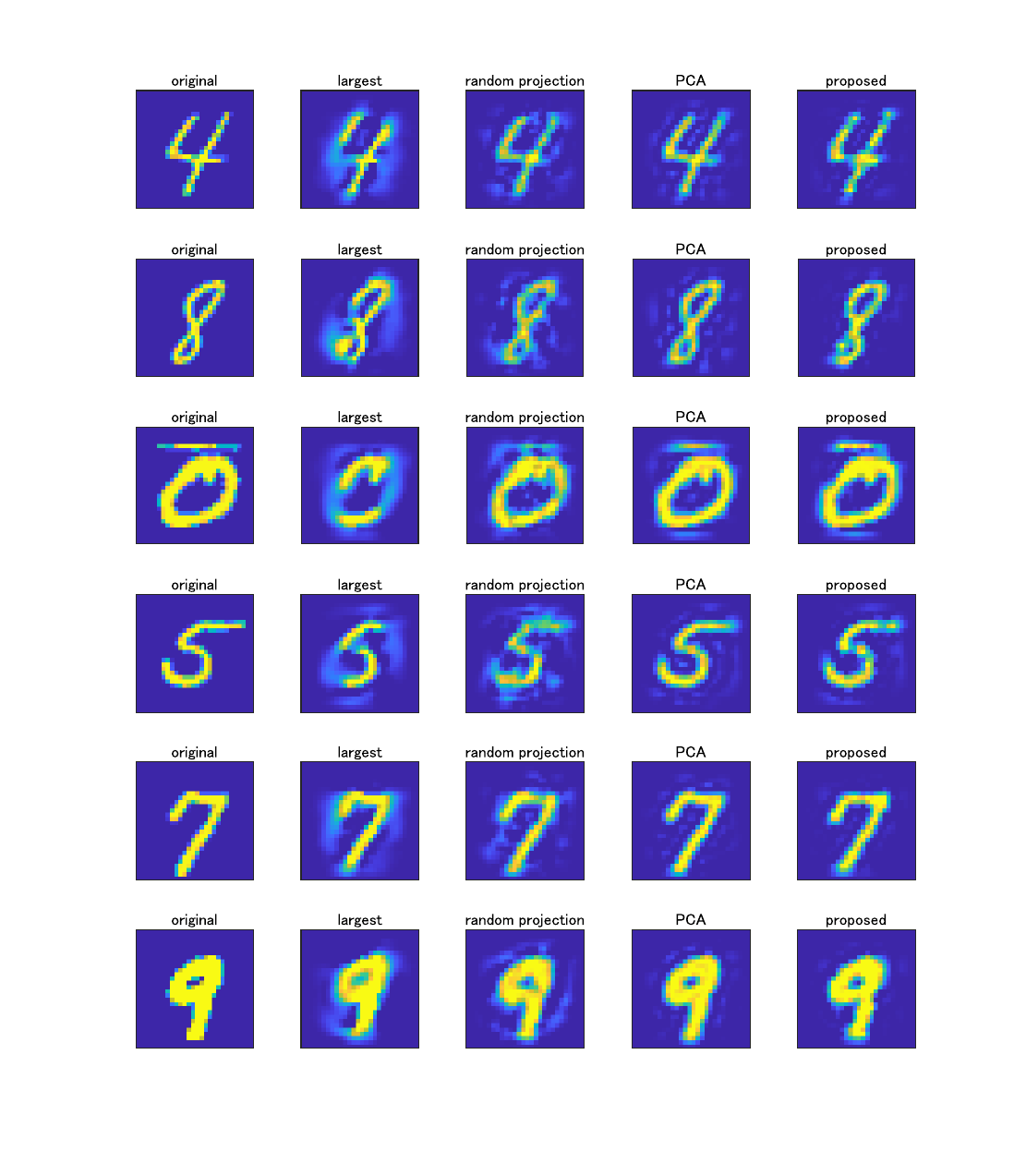}
  \vspace{-23mm}
  \caption{Examples of reconstruction from low-dimensional vectors.
  From left: original, variance-based selection, random projection, PCA, and the proposed selection.}
  \label{fig:reconst}
\end{figure}

Figure~\ref{fig:reconst} shows examples of reconstruction. From each low-dimensional $\B{y}$, $\B{x}$ was estimated using \eqref{eq:Q0} with $\B{z}=\B{x}$, the mean added back, and intensities clipped to $[0,255]$. We compare variance-based selection, random projection~\cite{Achlioptas2003}, PCA, and the proposed selection (all 100-D).

\section{Conclusion}
We proposed a swap-based element selection method for multiplication-free dimension reduction, where the selected subset is designed to minimize the mean-squared error of linear regression from the selected elements to a target vector (with reconstruction as a special case).
To make the discrete local search practical, we derived a fast evaluation of candidate swaps using the matrix inversion lemma, which reduces the per-candidate cost to inexpensive operations involving a $2\times 2$ matrix.
Preliminary experiments on handwritten digits demonstrated that the proposed acceleration enables effective element selection with reasonable computation time.
Future work includes broader application-oriented evaluations and extensions of the fast evaluation to other selection criteria.

\paragraph*{Acknowledgments}
This work was supported by JST CREST (JPMJCR19A3). The research motivation arose from discussions with Masahiro Sunohara and Chiho Haruta at RION Co., Ltd., to whom we express our gratitude.

\end{document}